\documentclass[runningheads]{llncs}

\usepackage[mobile]{eccv}

\usepackage{eccvabbrv}

\usepackage{graphicx}
\usepackage{booktabs}
\usepackage{float}
\usepackage[accsupp]{axessibility}  %
\usepackage{url}
\usepackage{multirow}
\usepackage{booktabs}
\usepackage{pifont}
\usepackage{xcolor}
\usepackage{graphicx}
\usepackage{amsmath}
\usepackage{algorithm}
\usepackage{mathtools}
\usepackage{algorithm}
\usepackage{algpseudocode}
\usepackage{listings}
\usepackage{wrapfig}
\usepackage{subcaption}
\usepackage{colortbl}
\usepackage{array}

\newcommand{\cmark}{\color{green}\ding{51}}%
\newcommand{\xmark}{\color{red}\ding{55}}%

\newcommand{\Mat}{\boldsymbol}
\newcommand{\Set}{\mathcal}

\newcommand{\real}{\mathbb{R}}

\newcommand{\ord}[1]{#1^{\rm{th}}}
\newcommand{\paragraphVspace}

\definecolor{top1}{RGB}{255,179,179}
\definecolor{top2}{RGB}{255,217,179}
\definecolor{top3}{RGB}{255,255,179}

\usepackage{hyperref}

\usepackage{orcidlink}

\begin{document}

\title{FSGS: Real-Time Few-shot View Synthesis \\ using Gaussian Splatting}

\author{%
  Zehao Zhu\inst{1\dagger}\index{Zehao, Zhu}, 
  Zhiwen Fan\inst{1\dagger}\index{Zhiwen, Fan}, 
  Yifan Jiang\inst{1}\index{Yifan, Jiang}, 
  Zhangyang Wang\inst{1}\index{Zhangyang, Wang}
}

\institute{
 ${}^{\dagger}$Equal Contribution, The University of Texas at Austin
}
\maketitle

\begin{abstract}

Novel view synthesis from limited observations remains a crucial and ongoing challenge.
 In the realm of NeRF-based few-shot view synthesis, there is often a trade-off between the accuracy of the synthesized view and the efficiency of the 3D representation.
To tackle this dilemma, we introduce a \textbf{\underline{F}}ew-\textbf{\underline{S}}hot view synthesis framework based on 3D \textbf{\underline{G}}aussian \textbf{\underline{S}}platting, which facilitates real-time, photo-realistic synthesis from a minimal number of training views.
\textbf{FSGS} employs an innovative \textit{Proximity-guided Gaussian Unpooling}, specifically designed for sparse-view settings, to bridge the gap presented by the extremely sparse initial point sets. This method involves the strategic placement of new Gaussians between existing ones, guided by a Gaussian proximity score, enhancing the adaptive density control.
We have identified that Gaussian optimization can sometimes result in overly smooth textures and a propensity for overfitting when training views are limited.  To mitigate these issues, FSGS introduces the synthesis of virtual views to replicate the parallax effect experienced during training, coupled with geometric regularization applied across both actual training and synthesized viewpoints. This strategy ensures that new Gaussians are placed in the most representative locations, fostering more accurate and detailed scene reconstruction.
Our comprehensive evaluation across various datasets—including NeRF-Synthetic, LLFF, Shiny, and Mip-NeRF360 datasets—illustrates that FSGS not only delivers exceptional rendering quality but also achieves an inference speed more than 2000 times faster than existing state-of-the-art methods for sparse-view synthesis. Project webpage \url{https://zehaozhu.github.io/FSGS/}.

\end{abstract}

\section{Introduction}
\label{sec:intro}
Novel view synthesis (NVS) from a set of view collections, as demonstrated by recent works~\cite{tewari2022advances, rabby2023beyondpixels, gao2023nerf}, has played a critical role in the domain of 3D vision and is pivotal in many applications, e.g., VR/AR and autonomous driving. Despite its effectiveness in photo-realistic rendering, the requirement of dense support views has hindered its practical usages~\cite{niemeyer2022regnerf}. Previous studies have focused on reducing the view requirements by leveraging Neural Radiance Field (NeRF)\cite{mildenhall2021nerf}, a powerful implicit 3D representation that captures scene details, combined with volume rendering techniques\cite{drebin1988volume}. Depth regularization~\cite{niemeyer2022regnerf,xu2022sinnerf,deng2023nerdi,truong2023sparf} within the density field, additional supervision from 2D pre-trained models~\cite{wang2023sparsenerf,jain2021putting}, large-scale pre-training~\cite{chen2021mvsnerf,yu2021pixelnerf}, and frequency annealings~\cite{yang2023freenerf} have been proposed and adopted to address the challenge of few-shot view synthesis.
While these NeRF-based approaches are promising, they often lead to substantial computational demands, which can affect real-time performance adversely. Subsequent research has managed to reduce training time in real-world scenarios from days to mere hours~\cite{sun2022direct,wang2022fourier,schwarz2022voxgraf,yu2021plenoxels}, and even minutes in some cases~\cite{muller2022instant}. However, a noticeable gap persists between attaining real-time rendering speeds and the desired photo-realistic, high-resolution output quality.

\begin{figure}[t!]
\centering
  \includegraphics[width=0.83\textwidth]{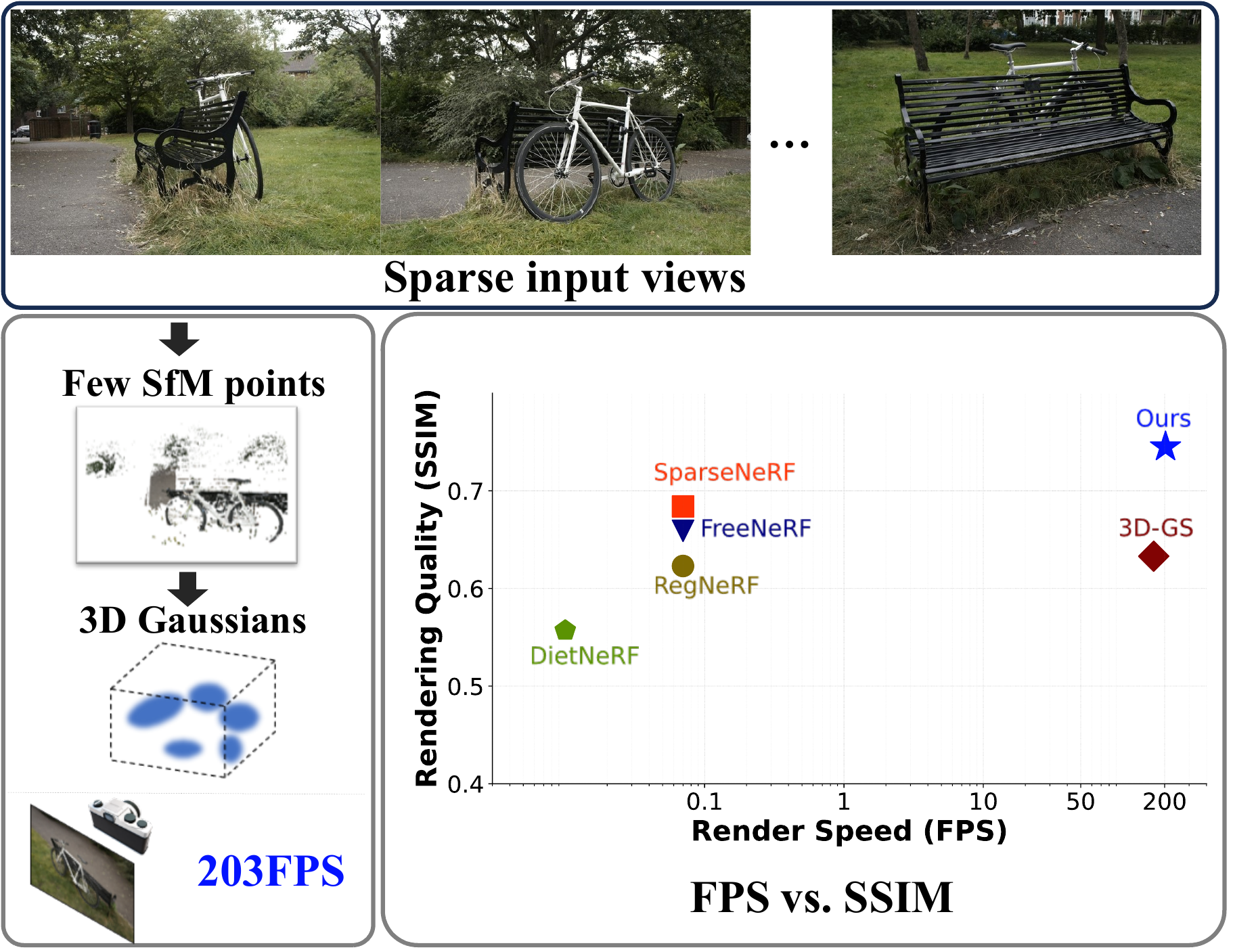}
  \vspace{-3mm}
  \caption{\textbf{Real-Time Few-shot Novel View Synthesis.} We present a point-based framework that is initialized from extremely sparse SfM points, achieving a significantly faster rendering speed (2900$\times$) while enhancing the visual quality (from 0.684 to 0.745, in SSIM) compared to the previous SparseNeRF~\cite{wang2023sparsenerf}.}
  \label{fig:teaser}
\end{figure}

In our research, we delve into the advancements in efficient 3D Gaussian Splatting (3D-GS)~\cite{kerbl20233d} and examine the challenges associated with deploying 3D-GS on sparse inputs. A crucial aspect for the effectiveness of 3D-GS is the densification process, which transforms the sparse initial point cloud into a more detailed representation of the 3D environment. However, the placement of new Gaussians, dictated by the spatial gradient, tends to be noisy and unrepresentative, especially in sparse-view scenarios. Additionally, the reliance on photometric loss with limited view counts often results in overly smooth textures when adhering to the conventional densification approach.

To address these issues, we introduce \textit{Proximity-guided Gaussian Unpooling}, a novel strategy designed specifically for sparse inputs. This method enhances the Gaussian representation by inserting new Gaussians between existing ones, based on the proximity to their neighbors. This strategic placement, combined with the initialization using observations from existing Gaussians, significantly improves scene representation by increasing the Gaussian density. Furthermore, we advocate for view augmentation through the generation of virtual camera, not present during training, to apply additional constraints in sparse setups. Incorporating monocular depth priors helps regularize the new Gaussians, steering them towards a plausible solution while enhancing texture detail, essential for accurate relative positioning in both actual training and synthetic camera views.
We have conducted thorough evaluations of our Few-Shot Gaussian Splatting (FSGS) framework across a variety of few-shot Novel View Synthesis benchmarks. These include the object-centric NeRF-Synthetic datasets, the forward-facing LLFF datasets, the Shiny datasets with intricate lighting conditions, and the unbounded Mip-NeRF360 datasets. Our experiments demonstrate that FSGS sets a new benchmark in rendering quality and operates at a real-time speed (203 FPS), making it suitable for real-world applications. The efficacy of our method allows FSGS to outperform 3D-GS even with fewer Gaussians, enhancing both the efficiency and quality on the rendered scenes.

Our key contributions are as follows:
\begin{itemize}
\item We propose a novel point-based framework, FSGS, for few-shot view synthesis that densifies new Gaussians via Proximity-guided Gaussian Unpooling. This method effectively increases the density of Gaussians, ensuring detailed and comprehensive scene representation.
\item FSGS addresses the overfitting challenge inherent in sparse-view Gaussian splatting. It achieves this by generating unseen viewpoints during training and incorporating distance correspondences on both training and synthesized pseudo views. This strategy directs the Gaussian optimization process toward solutions that are both highly accurate and visually compelling.
\item FSGS significantly enhances the visual quality, and also facilitates real-time rendering speeds (over 200 FPS) leading to a viable option for practical implementation in various real-world applications. 
\end{itemize}

\section{Related Works}

\subsection{Neural Representations for 3D Reconstruction}

The recent advancement of neural rendering techniques, such as Neural Radiance Fields (NeRFs)~\cite{mildenhall2021nerf}, has shown encouraging progress for novel view synthesis.
NeRF learns an implicit neural scene representation that utilizes a MLP to map 3D coordinates $(x, y, z)$ and view dependency $(\theta, \phi)$ to color and density through a volume rendering function. 
Tremendous works focus on improving its efficiency~\cite{fridovich2022plenoxels,muller2022instant,reiser2021kilonerf,sun2022direct,kerbl20233d, chen2022tensorf, garbin2021fastnerf}, quality~\cite{verbin2021ref,barron2021mip,barron2022mip,guo2022nerfren,suhail2022light,chen2022aug, wang2023f2nerf, barron2023zipnerf}, generalizing to unseen scenes~\cite{yu2021pixelnerf,chen2021mvsnerf,wang2021ibrnet, johari2022geonerf, t2023is, SRF}, applying artistic effects~\cite{fan2022unified,wang2022clip,jain2022zero, zhang2022arf} and 3D generation~\cite{poole2022dreamfusion, magic3d, liu2023zero1to3, karnewar2023holodiffusion, tang2023makeit3d, seo2023let, hoellein2023text2room, cao2023dreamavatar, gu2021stylenerf,chanmonteiro2020pi-GAN}. In particular, Reiser \textit{et al.}~\cite{reiser2021kilonerf} accelerate NeRF's training by splitting a big MLP into thousands of tiny MLPs. MVSNeRF~\cite{chen2021mvsnerf} constructs a 3D cost volume~\cite{yao2018mvsnet,gu2020cascade} and renders high-quality images from novel viewpoints. Moreover, Mip-NeRF~\cite{barron2021mip} adopts conical frustum rather than a single ray in order to mitigate aliasing. Mip-NeRF 360~\cite{Barron2022MipNeRF3U} further extends it to the unbounded scenes. While these NeRF-like models present strong performance on various benchmarks, they generally require several hours of training time. Muller \textit{et al.}~\cite{muller2022instant} adopt a multiresolution hash encoding technique that reduces the training time significantly. Kerbl \textit{et al.}~\cite{kerbl20233d} propose to use a 3D Gaussian Splatting pipeline that achieves real-time rendering for either objects or unbounded scenes. The proposed FSGS approach is based on the 3D Gaussian Splatting framework but largely reduces the required training views.

\subsection{Novel-View Synthesis Using Sparse Views}
The original neural radiance field takes more than one hundred images as input, largely prohibiting its practical usage. To tackle this issue, several works have attempted to reduce the number of training views.
Specifically, Depth-NeRF~\cite{deng2021depth} applies additional depth supervision to improve the rendering quality. RegNeRF~\cite{niemeyer2021regnerf} proposes a depth smoothness loss as geometry regularization to stabilize training. 
DietNeRF~\cite{jain2021putting} adds supervision on  the CLIP embedding space~\cite{radford2021learning}, to constraint the rendered unseen views.
PixelNeRF~\cite{yu2021pixelnerf} trains a convolution encoder to capture context information and learns to predict 3D representation from sparse inputs. More recently, FreeNeRF~\cite{yang2023freenerf} proposes a dynamic frequency controlling module for few-shot NeRF. SparseNeRF~\cite{wang2023sparsenerf} proposes a new spatial continuity loss to distill spatial coherence from monocular depth estimators. 
Concurrent work ReconFusion~\cite{wu2023reconfusion} employs diffusion models to synthesize additional views, which may not always adhere to view consistency and time consuming. ReconFusion jointly train a Zip-NeRF with synthesized views under a sparse-view setting. In contrast, our method improves the optimization process of Gaussian Splatting, and facilitates both the real-time rendering speed and rendering quality.

\section{Method}

\begin{figure*}[t]
\vspace{-1mm}
\vskip 0.2in
\begin{center}
    \centering
    \includegraphics[width=1.0\linewidth]{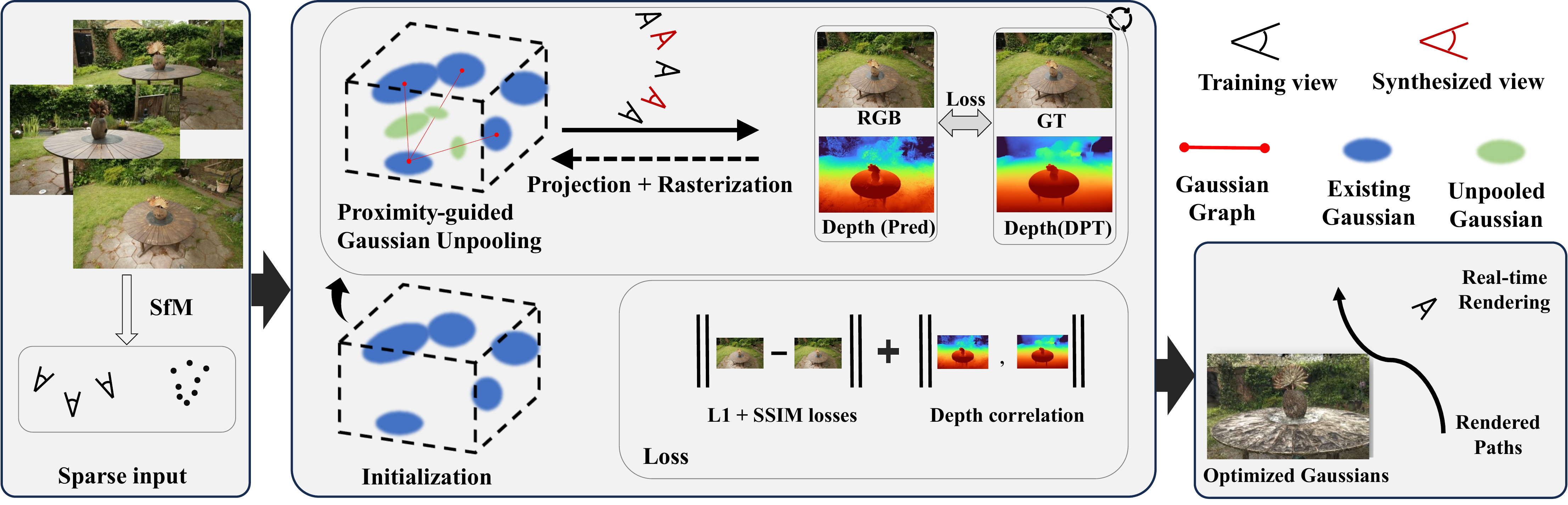}
\vspace{-7mm}
\caption{\textbf{FSGS Pipeline.} 3D Gaussians are initialized from COLMAP, with a few images (\textbf{black cameras}). For the sparsely placed Gaussians, we propose densifying new Gaussians to enhance scene coverage by unpooling existing Gaussians into new ones, with properly initialized Gaussian attributes. Monocular depth priors, enhanced by sampling unobserved views (\textcolor{red}{\textbf{red cameras}}), guide the optimization of grown Gaussians towards a reasonable geometry. The final loss consists of a photometric loss term, and a geometric regularization term calculated as depth relative correspondence.}
\label{fig:arc}
\end{center}
\vspace{-6mm}
\end{figure*}

\paragraph{\textbf{Overview.}} 
This section provides an overview of the FSGS framework, as illustrated in Fig.~\ref{fig:arc}. FSGS processes a limited set of images captured from a static scene. The camera poses and point clouds are derived using the Structure-from-Motion (SfM) software, COLMAP\cite{schonberger2016structure}. The initialization of 3D Gaussians is based on a sparse point cloud, incorporating attributes such as color, position, and a predefined conversion rule for shape and opacity.
The issue of extremely sparse points is tackled through the implementation of Proximity-guided Gaussian Unpooling. This method densifies Gaussians and populates the empty spaces by assessing the proximity between existing Gaussians and positioning new ones in the most representative areas, thereby enhancing scene details.
To mitigate overfitting in standard 3D-GS with sparse-view data, we introduce the generation of pseudo camera viewpoints around the training cameras. This approach, coupled with the geometry regularization, steers the model towards accurately reconstructing the scene's geometry.

\subsection{Preliminary and Problem Formulation}
3D Gaussian Splatting (3D-GS), as delineated in Kerbl et al.~\cite{kerbl20233d}, represents an 3D scene explicitly through a collection of 3D Gaussians, with attributes: a position vector $\Mat \mu \in \mathbb{R}^3$ and a covariance matrix $ \Sigma\in \mathbb{R}^{3\times 3}$. Each Gaussian influences a point $\Mat x$ in 3D space following the 3D Gaussian distribution:
\vspace{-2mm}
\begin{equation}
\label{eq:3dgs}
    G(\Mat x)=\frac{1}{\left(2\pi\right)^{3/2}\left|\Sigma\right|^{1/2}}e^{-\frac{1}{2}{\left(\Mat x-\Mat \mu\right)}^T\Sigma^{-1}\left(\Mat x-\Mat \mu\right)}
\end{equation}
\vspace{-3mm}

To ensure that $\Sigma$ is positive semi-definite and holds practical physical significance,  $ \Sigma$ is decomposed into two learnable components by $\Sigma = {R}{S}{S}^T{R}^T $,  where $ R$ is a quaternion matrix  representing rotation and $ S$ is a scaling matrix.

\begin{figure}[t]
\vspace{-2mm}
\begin{center}
    \centering
    \includegraphics[width=0.87\linewidth]{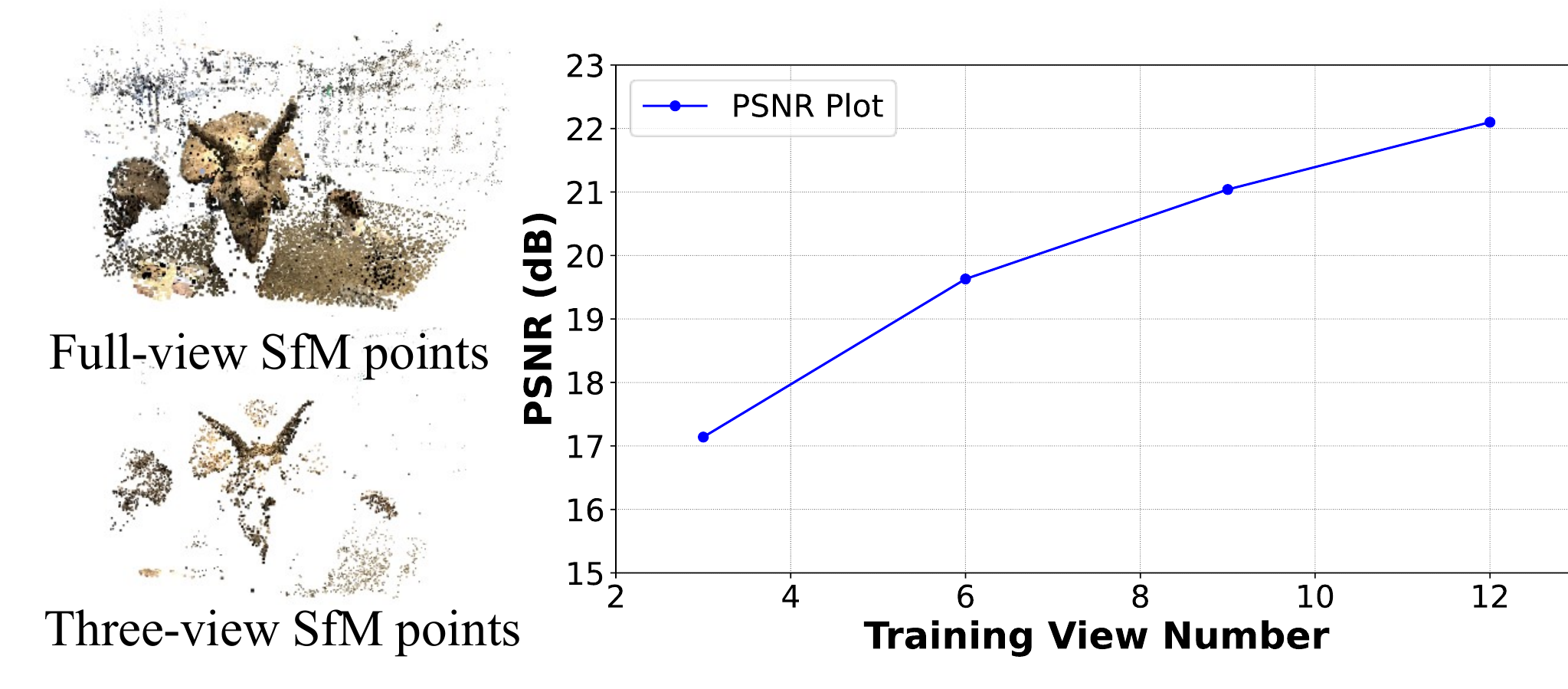}
\vspace{-5mm}
\caption{\textbf{Points Sparsity vs. Synthesized Quality.} The SfM points from COLMAP using 3-views (Bottom Left) is significantly sparse than full-view(Top Left). 3D-GS with sparse SfM points will decrease its quality when the training view number decreases.}\label{fig:3d-gs-difficulty}
\vspace{-6mm}
\end{center}
\end{figure}

Each Gaussians also store an opacity logit $ o \in \mathbb{R}$ and the appearance feature represented by $n$ spherical harmonic (SH) coefficients $\left\{ c_i\in \real^{3}|i=1,2,...,n\right\}$ where $n=D^2$ is the number of coefficients of SH with degree $D$. 
To render the 2D image, 3D-GS orders all the Gaussians that contributes to a pixel and blends the ordered Gaussians overlapping the pixels using the following function:
\vspace{-3.5mm}
\begin{equation}
\label{eq:render}
          c = \sum_{i=1}^n\ c_i \alpha_i  \prod_{j=1}^{i-1}(1-\alpha_j)
\end{equation}
\vspace{-5mm}

\noindent where $ c_i$ is the color computed from the SH coefficients of the $\ord i$ Gaussian. $\alpha_i$ is given by evaluating a 2D Gaussian with covariance $\Sigma^\prime \in \real^{2\times 2}$ multiplied by the opacity. 
The 2D covariance matrix $\Sigma^\prime$ is calculated by $\Sigma^\prime =  J  W \Sigma { W}^T { J}^T$, projecting the 3D covariance $\Sigma$ to the camera coordinates. Here, $ J$ denotes the Jacobian of the affine approximation of the projective transformation, $ W$ is the view transformation matrix. A heuristic Gaussian densification scheme is introduced in 3D-GS~\cite{kerbl20233d}, where Gaussians are densified based on an average magnitude of view-space position gradients which exceed a threshold.
Although this method is effective when initialized with comprehensive SfM points, it is insufficient for fully covering the entire scene with an extremely sparse point cloud, from sparse-view input images. Additionally, some Gaussians tend to grow towards extremely large volumes, leading to results that overfit the training views and generalize badly to novel viewpoints (See Fig.~\ref{fig:3d-gs-difficulty}).

However, 3D-GS is initialized from SfM points, and its performance strongly relies on both the quantity and accuracy of the initialized points. Although the subsequent Gaussian densification~\cite{kerbl20233d} can increase the number of Gaussians in both under-reconstructed and over-reconstructed regions, this straightforward strategy falls short in few-shot settings: it suffers from inadequate initialization, leading to oversmoothed outcomes and a tendency to overfit on training views.

\subsection{Proximity-guided Gaussian Unpooling}
\vspace{1mm}
The granularity of the modeled scene depends heavily on the quality of the 3D Gaussians representing the scene; therefore, addressing the limited 3D scene coverage is crucial for effective sparse-view modeling.
\vspace{-2mm}
\paragraph{\textbf{Proximity Score and Graph Construction.}}
During Gaussian optimization, we construct a directed graph, referred to as the proximity graph, to connect each existing Gaussian with its nearest $K$ neighbors by computing the proximity (a.k.a. Euclidean distance). 
Specifically, we denote the originating Gaussian at the head as the ``source'' Gaussian, while the one at the tail as the ``destination'' Gaussian, which is one of the source's $K$ neighbors.  These ``destination'' Gaussians are determined via the rule:
\vspace{-2mm}
\begin{align}
    D_i^K &= K\text{-min}(d_{ij}), \quad \forall j \neq i 
\end{align}

\vspace{-2mm}
Here, $d_{ij}$ is calculated via $d_{ij} = \| \mu_i - \mu_j \|$, representing the Euclidean distance  among the centers of Gaussian $G_i$ and Gaussian $G_j$.
The assigned proximity score $P_i$ to Gaussian $G_i$ is calculated as the average distance to its $K$ nearest neighbors:
\vspace{-5mm}
\begin{align}
    P_i &= \frac{1}{K} \sum_{j=1}^{K} D_i^K 
\end{align}

The proximity graph is updated following the densification or pruning process during optimization. We set $K$ to 3 in practice.

\begin{wrapfigure}[17]{b}{0.55\textwidth}
\centering
\vspace{-8mm}
  \includegraphics[width=0.56\textwidth]{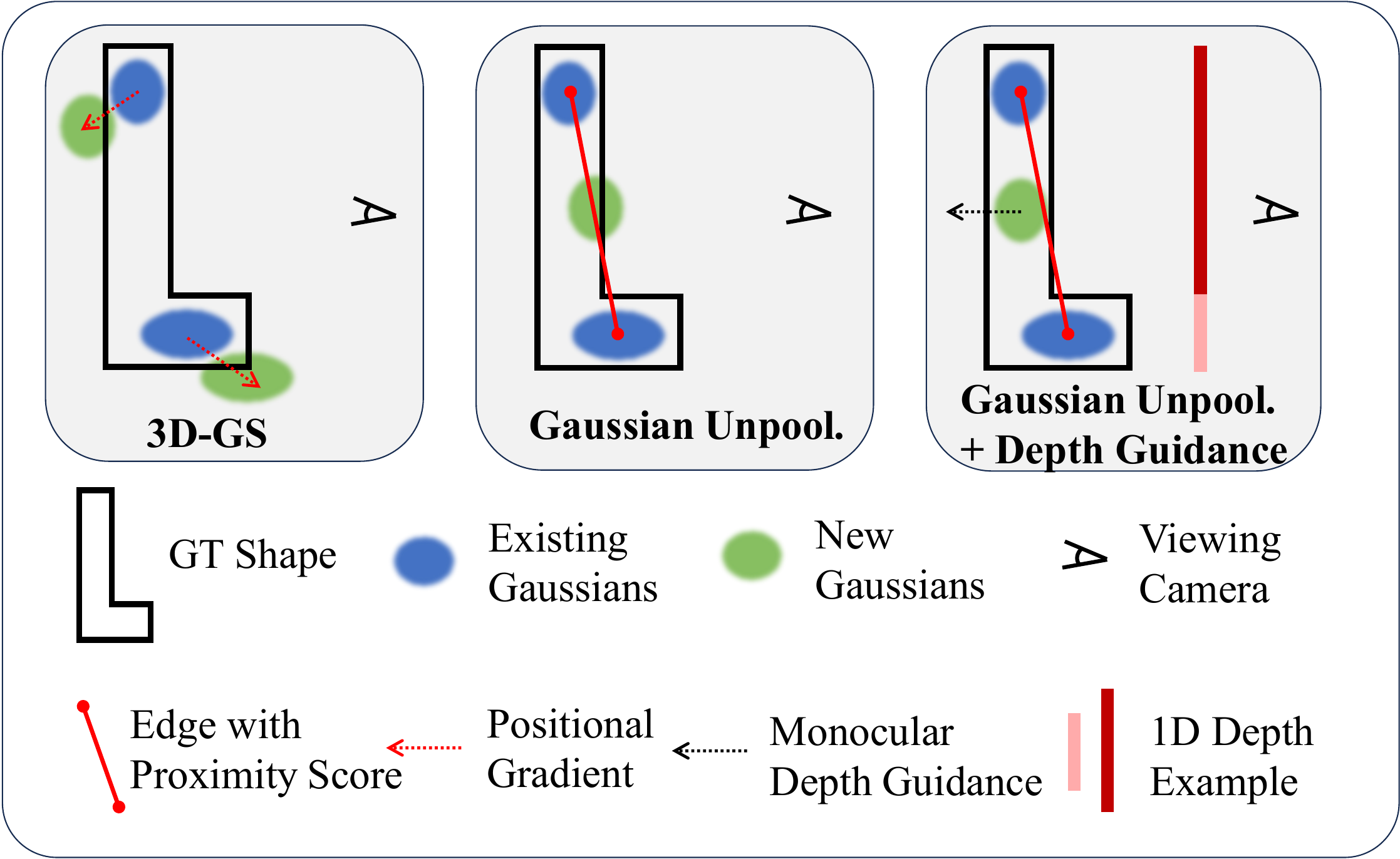}
    \vspace{-6.5mm}
  \caption{\textbf{Gaussian Unpooling Illustration.} We show a 2D toy case for visualizing Gaussian Unpooling with depth guidance, where the example 1D depth provides priors on the relative distance of the Gaussians from the viewing direction, guide the Gaussian deformation toward a better solution.}
  \label{fig:gs-unpool}
\end{wrapfigure}
\vspace{-3mm}
\paragraph{\textbf{Gaussian Unpooling.}}
Inspired by the vertex-adding strategy of the mesh subdivision algorithm~\cite{zorin1996interpolating} which is widely used in computer graphics, we propose unpooling Gaussians based on the proximity graph and the proximity score of each Gaussian. Specifically, if the proximity score of a Gaussian exceeds the threshold $t_\text{prox}$, our method will grow a new Gaussian at the center of each edge, connecting the ``source'' and ``destination'' Gaussians, as shown in Fig.~\ref{fig:gs-unpool}.
The attributes of scale and opacity in the newly created Gaussians are set to match those of the ``destination'' Gaussians. Meanwhile, other attributes such as rotation and SH coefficients are initialized to zero.
The Gaussian unpooling strategy encourages the newly densified Gaussians to be distributed around the representative locations and progressively fill observation gaps during optimization.

\vspace{-2mm}
\subsection{Geometry Guidance for Gaussian Optimization}
Having achieved dense coverage by unpooling Gaussians, a photometric loss with sparse-view clues is applied for optimizing Gaussians. 
However, the insufficient parallax in the sparse-view setting limit the 3D Gaussians to be optimized toward a globally consistent direction where it tend to overfit on training views, and poor generalization to novel views.
To inject more regularization to the optimization, we propose to create some virtual cameras that unseen in training, and apply the pixel wise geometric correspondences as additional regularization.

\vspace{-2mm}
\paragraph{\textbf{Synthesize Pseudo Views.}}
To address the inherent issue of overfitting to sparse training views, we employ unobserved (pseudo) view augmentation to incorporate more prior knowledge within the scene derived from a 2D prior model.
The synthesized view is sampled from the two closest training views in Euclidean space, calculating the averaged camera orientation and interpolating a virtual one between them.
A random noise is applied to the 3 degrees-of-freedom (3DoF) camera location as shown in
Eq.~\ref{eq:view_aug}, and then images are rendered.
\vspace{-2mm}
\begin{align}\label{eq:view_aug}
{ \Mat P}' = ({\Mat t} + {\varepsilon}, {\Mat q}), \quad {\varepsilon} \sim \mathcal{N}({0},{\delta})
\end{align}
\vspace{-6mm}

\noindent Here, ${\Mat t} \in {\Mat P}$ denotes camera location, while ${\Mat q}$ is a quaternion representing the rotation averaged from the two cameras.
This approach of synthesizing online pseudo-views enables dynamic geometry updates, as the 3D Gaussians will update progressively, reducing the risk of overfitting.

\vspace{-4mm}
\paragraph{\textbf{Inject Geometry Coherence from Monocular Depth.}}
We generate the monocular $\Mat D_{est}$ depth maps at both training and pseudo views by using the pre-trained Dense Prediction Transformer (DPT)~\cite{ranftl2021vision}, trained with 1.4 million image-depth pairs as a handy yet effective choice.
To mitigate the scale ambiguity between the true scene scale and the estimated depth, we introduce a relaxed relative loss, \textit{Pearson} correlation, on the estimated and rendered depth maps.
It measures the distribution difference between 2D depth maps and follows the below function:
\vspace{-4mm}
\begin{align}\label{eq:depthloss}
    \operatorname{Corr}(\hat{\Mat{D}}_{\text{ras}}, \hat{\Mat{D}}_{\text{est}}) = \frac{\operatorname{Cov}(\hat{\Mat{D}}_{\text{ras}}, \hat{\Mat{D}}_{\text{est}})} {\sqrt{\operatorname{Var}(\hat{\Mat{D}}_{\text{ras}})\operatorname{Var}(\hat{\Mat{D}}_{\text{est}})}}
\end{align}
\vspace{-4mm}

\noindent This soften constraint allows for the alignment of depth structure without being hindered by the inconsistencies in absolute depth values.
\vspace{-3mm}
\paragraph{\textbf{Differentiable Depth Rasterization.}}
To enable the backpropogation from depth prior to guide Gaussian training, we implement a differentiable depth rasterizor, allowing for receiving the error signal between the rendered depth $\Mat D_{ras}$ and the estimated depth $\Mat D_{est}$.
Specifically, we utilize the alpha-blending rendering in 3D-GS for depth rasterization, where the z-buffer from the ordered Gaussians contributing to a pixel is accumulated for producing the depth value:
\vspace{-6mm}
\begin{align}
      d = \sum_{i=1}^n d_i \alpha_i \prod_{j=1}^{i-1}(1-\alpha_j)
\end{align}
\vspace{-4mm}

\noindent Here $ d_i $ represents the z-buffer of the $\ord i$ Gaussians and $\alpha$ is identical to that in Eq.~\ref{eq:render}. 
This  differentiable implementation enables the depth correlation loss.

\vspace{-3mm}
\subsection{Optimization}
Combining all together, we can summarize the training loss:
\vspace{-1mm}
\begin{align} 
\mathcal{L}(\Mat{G}, \Mat{C}) &= \lambda_1 \underbrace{\left\lVert \mathbf{\Mat{C}} - \mathbf{\hat{\Mat{C}}} \right\rVert_1}_{\mathcal{L}_{\text{1}}} + 
\lambda_2 \underbrace{\operatorname{D-SSIM}({\Mat{C}}, \hat{\Mat{C}})}_{\mathcal{L}_{\text{ssim}}} + \lambda_3 \underbrace{\left\lVert \operatorname{Corr}(\Mat{D}_{\text{ras}}, \Mat{D}_{\text{est}}) \right\rVert_1}_{\mathcal{L}_{\text{regularization}}}
\label{eqn:final_loss}
\end{align}

\vspace{-3mm}
\noindent where $\mathcal{L}_{\text{1}}$, and $\mathcal{L}_{\text{ssim}}$ stands for the photometric loss term between predicted image $\hat{\Mat{C}}$ and ground-truth image $\Mat{C}$. $\mathcal{L}_{\text{regularization}}$ represents the geometric regularization term on both the training views and synthesized pseudo views.
We set $\lambda_1$, $\lambda_2$, $\lambda_3$ as 0.8, 0.2, 0.05 respectively by grid search. 
The pseudo views sampling is enabled after 2,000 iterations to ensure the Gaussians can roughly represent the scene.

\vspace{-1mm}
\section{Experiments}
\vspace{-1mm}
\subsection{Experimental Settings}
\textbf{LLFF Datasets~\cite{mildenhall2019local}} consist of  eight forward-facing real-world scenes. Following RegNeRF~\cite{niemeyer2022regnerf}, we select every eighth image as the test set, and evenly sample sparse views from the remaining images for training. We utilize 3 views to train all the methods, and downsample their resolutions to 4$\times$ and 8$\times$, which are $504 \times 378$ and $1008 \times 756$ respectively. 

\noindent \textbf{Mip-NeRF360 Datasets~\cite{Barron2022MipNeRF3U}} consist of nine scenes,  each featuring a complex central object or area against a detailed background. 
We utilize 24 training views for comparison, with images downsampled to 4$\times$ and 8$\times$. Test images are selected the same with LLFF Datasets. We aim to establish this challenge benchmark for testing few-shot view synthesis in complex outdoor scenarios. 

\noindent \textbf{NeRF-Synthetic Datasets (Blender)~\cite{mildenhall2021nerf}} have eight objects with realistic images synthesized by Blender.
We align with DietNeRF~\cite{jain2021putting}, where we use 8 images for training and 25 for testing, at resolution of 400$\times$400.  

\noindent  \textbf{Shiny Datasets~\cite{shiny}} contain more challenging view-dependent effects, like the rainbow reflections on a CD and refraction through a liquid bottle. We evenly select 3 views from the Shiny datasets for training at resolutions of 504×378.

\vspace{-3mm}
\paragraph{\textbf{Baselines.}}
We compare FSGS with several few-shot NVS methods on these three dataset, including DietNeRF~\cite{jain2021putting}, RegNeRF~\cite{niemeyer2022regnerf}, FreeNeRF~\cite{yang2023freenerf}, and SparseNeRF~\cite{wang2023sparsenerf}. Additionally, we include comparisons with the high-performing Mip-NeRF~\cite{barron2022mip}, primarily designed for dense-view training, and point-based 3D-GS, following its original dense-view training recipe. Following ~\cite{niemeyer2022regnerf, wang2023sparsenerf, kerbl20233d}, we report the average PSNR, SSIM, LPIPS scores and FPS for all the methods.

\vspace{-3mm}
\paragraph{\textbf{Implementation Details.}}
We implemented FSGS using the PyTorch framework, with initial point cloud computed from SfM, using only the training views.
During optimization, we densify the Gaussians every 100 iterations and start densification after 500 iterations. 
The total optimization steps are set to 10,000, requiring approximately 9.5 minutes on LLFF datasets, and  $\sim$24 minutes on Mip-NeRF360 datasets.
We set proximity threshold $t_\text{prox}$ to 10, and the pseudo views are sampled after 2,000 iterations. 
We utilize the pre-trained DPT model~\cite{ranftl2021vision} for depth estimation.  All results are obtained using a NVIDIA A6000 GPU.

\subsection{Comparisons to other Few-shot Methods}
\paragraph{\textbf{Comparisons on LLFF Datasets.}}
As shown in Tab.~\ref{tab:main1}, our method FSGS, despite trained from sparse SfM point clouds, provides the best quantitative results and effectively addresses the insufficient scene coverage in the initialization. 
Our method surpasses  SparseNeRF by 0.45dB and 0.81dB  in PSNR at both test resolutions, while inferencing 2180 times faster, which makes FSGS a viable choice for practical usages. 
FSGS also outperforms 3D-GS by 2.88dB in PSNR and boost the FPS from 385 to 458, demonstrating that our refined Gaussians are more compact for scene representation from sparse views. 

\begin{figure}[!]
\vspace{-1mm}
\begin{center}
    \centering
    \includegraphics[width=0.99\linewidth]{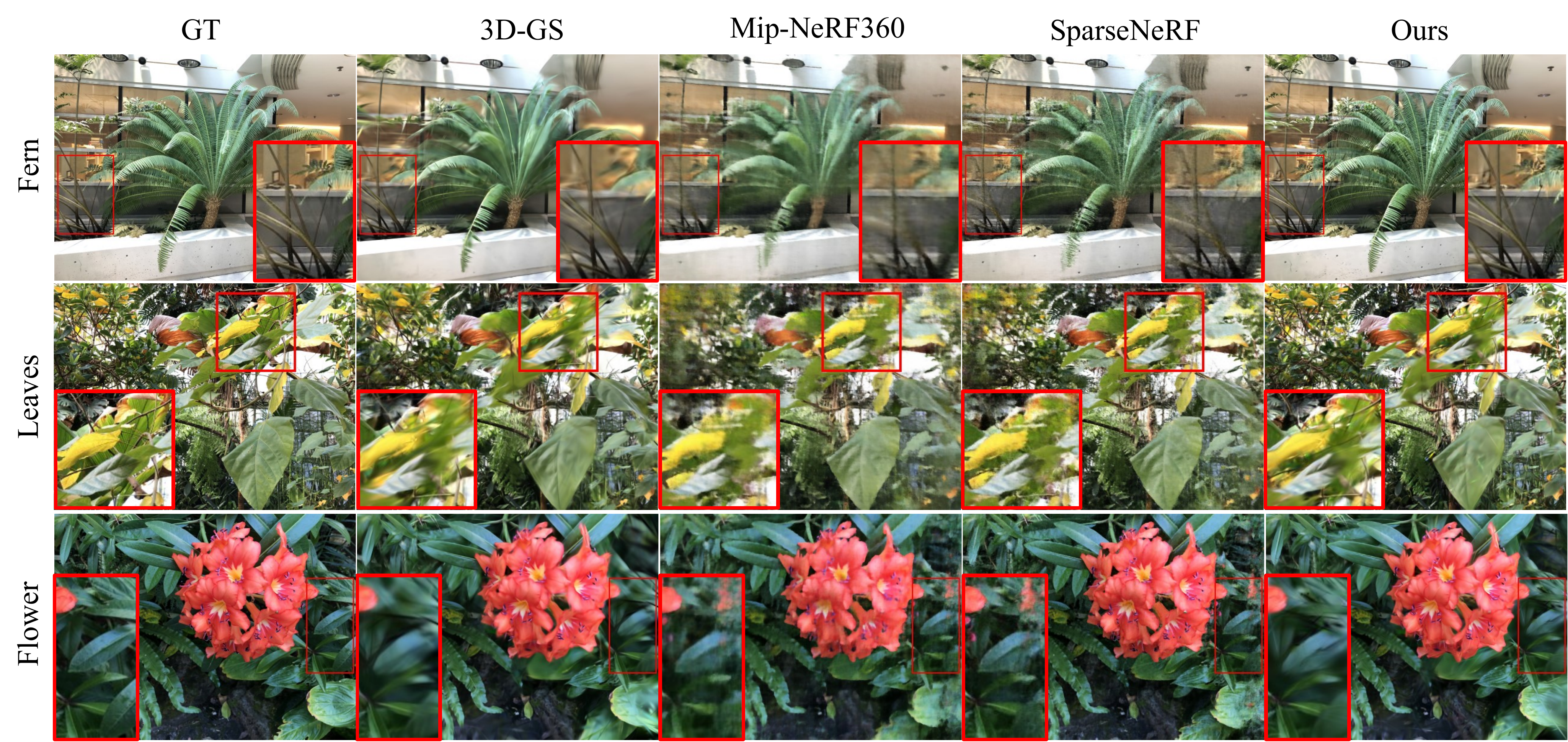}
\vspace{-3mm}
\caption{\textbf{Qualitative Results on LLFF Datasets.} We demonstrate novel view results produced by 3D-GS~\cite{kerbl20233d}, Mip-NeRF360~\cite{Barron2022MipNeRF3U}, SparseNeRF~\cite{wang2023sparsenerf} and our approach for comparison. We can observe that NeRF-based methods generate floaters (Scene: \textit{Flower}) and show aliasing results (Scene: \textit{Leaves}) due to limited observation. 3D-GS produces oversmoothed results,  caused by overfitting on training views. Our method produces pleasing appearances while demonstrating detailed thin structures. }
\label{fig:exp_llff}
\end{center}
\vspace{-6mm}
\end{figure}

The qualitative analysis, as presented in Fig.~\ref{fig:exp_llff}, demonstrates that Mip-NeRF and 3D-GS struggle with the extreme sparse view problem; Mip-NeRF~\cite{barron2021mip} leads to degraded geometric modeling, and 3D-GS produces blurred results in areas with complex geometry.
The geometry regularization in RegNeRF~\cite{niemeyer2021regnerf}, SparseNeRF~\cite{wang2023sparsenerf} and frequency annealing in FreeNeRF~\cite{yang2023freenerf} do improve the quality to some extent, but still exhibit insufficient visual quality.
In contrast, our proposed \textit{Proximity-Guided Gaussian Unpooling}, and the relative geometric regularizations on both training and synthesized virtual views, pulls more Gaussians to the unobserved regions, and thus recovers more textural and structural details.

\begin{table}[t]
\centering
\caption{\textbf{Quantitative Comparison in LLFF Datasets, with 3 Training Views.} FSGS achieves the best performance in terms of rendering accuracy and inference speed across all resolutions.
Significantly, FSGS runs 2,180$\times$ faster than the previous best, SparseNeRF, while improving the SSIM from 0.624 to 0.652, at the resolution of $504 \times 378$.
We color each cell as \textbf{\colorbox[RGB]{255,179,179}{best}}, \textbf{\colorbox[RGB]{255,217,179}{second best}},
and \textbf{\colorbox[RGB]{255,255,179}{third best}}. }
\vspace{-4mm}
\begin{tabular}{c|cccc|cccc} \toprule
\multirow{2}{*}{Methods} & \multicolumn{4}{c|}{1/8 Resolution}                                                & \multicolumn{4}{c}{1/4 Resolution}                                                  \\ \cline{2-9}
                                  & FPS$\uparrow$ & PSNR$\uparrow$ & SSIM$\uparrow$ & LPIPS$\downarrow$ & FPS$\uparrow$ & PSNR$\uparrow$ & SSIM$\uparrow$ & LPIPS$\downarrow$   \\ \hline
Mip-NeRF                          & \cellcolor{top3} 0.21             & 16.11              & 0.401              & 0.460                               & \cellcolor{top3} 0.14            &  15.22              &       0.351         &     0.540              \\
3D-GS                             & \cellcolor{top2}385             & 17.43              & 0.522              & \cellcolor{top3}0.321                               &       \cellcolor{top2}312       &    16.94            &        0.488        &      0.402             \\ \hline
DietNeRF                          & 0.14             & 14.94              & 0.370               & 0.496                             & 0.08             &    13.86            &       0.305         &       0.578            \\
RegNeRF                           & \cellcolor{top3}0.21             & 19.08              & 0.587              & 0.336                              & \cellcolor{top3}0.14            &    18.06            &         0.535       &      0.411             \\
FreeNeRF                  & \cellcolor{top3}0.21         & \cellcolor{top3}19.63             & \cellcolor{top3}0.612              & \cellcolor{top2}0.308                                         & \cellcolor{top3}0.14             &   \cellcolor{top3}18.73             &     \cellcolor{top3}0.562           &     \cellcolor{top2} 0.384             \\
SparseNeRF         & \cellcolor{top3}0.21                 & \cellcolor{top2}19.86             & \cellcolor{top2}0.624               & 0.328                                           &     \cellcolor{top3}0.14          &    \cellcolor{top2}19.07            &    \cellcolor{top2} 0.564           &      \cellcolor{top3}0.401             \\

\textbf{Ours}                              & \cellcolor{top1}{458}            & \cellcolor{top1}{20.31}             & \cellcolor{top1}{0.652}              & \cellcolor{top1}{0.288}                                & \cellcolor{top1}{351}            &    \cellcolor{top1}{19.88}            &      \cellcolor{top1}{0.612}          &       \cellcolor{top1}{0.340}                             \\ 
\bottomrule
\end{tabular}
\label{tab:main1}
\end{table}

It is worth to note that, the rendering speed of the 3D Gaussian representation is contingent upon the number of optimized Gaussians, where a reduced Gaussian count leads to faster rendering speed. FSGS outperforms 3D-GS while using less Gaussian counts, by effectively places and optimizes the Gaussians to the most representative positions than 3D-GS.
With the same initialization method as 3D-GS, the average optimized Gaussian count from FSGS is \textbf{57,513} on LLFF dataset, considerably lower than \textbf{63,219} from 3D-GS, which results in a faster rendering speed of FSGS than 3D-GS.

\begin{table}[t]
\caption{\textbf{Quantitative Comparison in Mip-NeRF360 Datasets, with 24 Training Views.} Our FSGS shows obvious advantages over NeRF-based methods, with an improvement of more than 0.05 in SSIM and running 4,142$\times$ faster. Additionally, our method not only performs better than 3D-GS in rendering metrics but also shows improvement in FPS (from 223 to 290), thanks to the Gaussian unpooling which motivates Gaussians to expand to unseen regions more accurately. }
\vspace{-4mm}
\centering
\begin{tabular}{c|cccc|cccc} \toprule
\multirow{2}{*}{\textbf{Methods}} & \multicolumn{4}{c|}{1/8 Resolution}                                                & \multicolumn{4}{c}{1/4 Resolution}                                                  \\ \cline{2-9}
                                  & FPS$\uparrow$ & PSNR$\uparrow$ & SSIM$\uparrow$ & LPIPS$\downarrow$ & FPS$\uparrow$ & PSNR$\uparrow$ & SSIM$\uparrow$ & LPIPS$\downarrow$ \\ \hline
Mip-NeRF360                      &  \cellcolor{top3}0.12             & 21.23             & 0.613              & 0.351                               & \cellcolor{top3}0.07             &     19.78            &     0.530           &     0.431             \\
3D-GS                             & \cellcolor{top2}223             & 20.89              & 0.633             & \cellcolor{top3}0.317                               &     \cellcolor{top2}145          &   19.93             &     \cellcolor{top3}0.588           &      0.401            \\ \hline
DietNeRF                          & 0.05             & 20.21              & 0.557              & 0.387                               & 0.03             &       19.11         &    0.482            &       0.452            \\
RegNeRF                           & 0.07             & 22.19              & 0.643              & 0.335             & 0.04             &        20.55        &   0.546             &      0.398            \\
FreeNeRF                          & 0.07             & \cellcolor{top3}22.78              & \cellcolor{top3}0.689              & 0.323   &     0.04                & \cellcolor{top3}21.04           &      0.587          &    \cellcolor{top2}0.377               \\
SparseNeRF                        & 0.07             & \cellcolor{top2}22.85              & \cellcolor{top2}0.693              & \cellcolor{top2}0.315              &       0.04        &     \cellcolor{top2}21.13           &     \cellcolor{top2}0.600           &     \cellcolor{top3}0.389              \\

Ours                              & \cellcolor{top1}290             & \cellcolor{top1}23.70              &  \cellcolor{top1}0.745             &   \cellcolor{top1}0.220                           & \cellcolor{top1}203             &  \cellcolor{top1}22.82              &         \cellcolor{top1}0.693      &       \cellcolor{top1}0.293            \\  \bottomrule
\end{tabular}
\label{tab:main2}
\end{table}

\begin{figure}[!]
\vspace{-1mm}
\begin{center}
    \centering
    \includegraphics[width=0.99\linewidth]{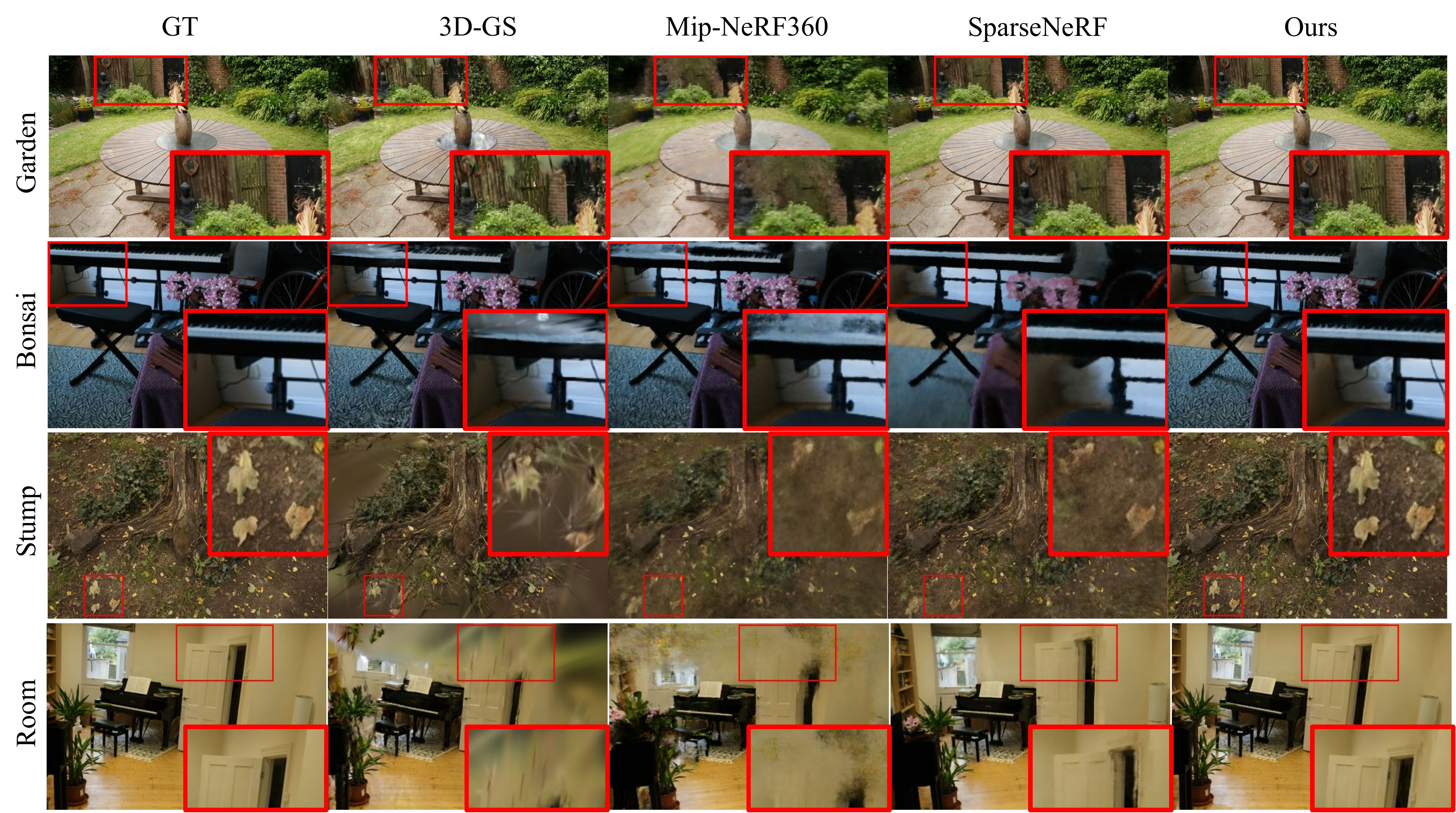}
    \vspace{-3mm}
\caption{\textbf{Qualitative Results on Mip-NeRF360 Datasets.}
Comparisons were conducted with 3D-GS~\cite{kerbl20233d}, Mip-NeRF360~\cite{Barron2022MipNeRF3U}, and SparseNeRF~\cite{wang2023sparsenerf}. Our method continues to produce visually pleasing results with sharper details than other methods in large-scale scenes.}
\label{fig:exp_360}
\end{center}
\vspace{-9mm}
\end{figure}

\vspace{-4mm}
\paragraph{\textbf{Comparisons on Mip-NeRF360 Datasets.}}
As shown in Tab.~\ref{tab:main2}, methods requiring dense view coverage (Mip-NeRF360, 3D-GS) are outperformed by ours in terms of rendering speed and metrics, across the two resolutions.
Methods employing regularizations from their respective geometry and appearance fields (DietNeRF, RegNeRF, FreeNeRF, SparseNeRF) still fall short in rendering quality, while remains far from achieving real-time speed.
Our FSGS significantly outperforms NeRF-based approaches, boosting PNSR by 0.85dB and improving FPS from 0.07 to 290 at 1/8 resolution.
We provide a qualitative comparison in Fig.~\ref{fig:exp_360}, where we observe that Mip-NeRF360 and SparseNeRF fail to capture the intricate details of scenes and tend to overfit on sparse training views, most notably in areas far away from cameras. 
In comparison, FSGS  recovers the fine-grained details such as the leaves on the ground (Scene: \textit{Stump}) and the piano keys (Scene: \textit{Bonsai}), aligning well with the ground truth.

\begin{table}[t]
\caption{\textbf{Quantitative Comparison in Blender and Shiny Datasets.} FSGS outperforms existing few-shot methods and 3D-GS across all metrics, validating the generalization of the proposed techniques to handheld object-level 3D modeling and datasets with challenging reflective effects. 
}
\vspace{-3mm}
\centering

\begin{tabular}{c|cccc|cccc} \toprule
\multirow{2}{*}{\textbf{Methods}} & \multicolumn{4}{c|}{Blender Dataset~\cite{mildenhall2021nerf}}                                                & \multicolumn{4}{c}{Shiny Dataset~\cite{shiny}}                                                  \\ \cline{2-9}
                                  & FPS$\uparrow$ & PSNR$\uparrow$ & SSIM$\uparrow$ & LPIPS$\downarrow$ & FPS$\uparrow$ & PSNR$\uparrow$ & SSIM$\uparrow$ & LPIPS$\downarrow$ \\ \hline
Mip-NeRF                             & \cellcolor{top3}0.22               & 20.89          & 0.830          & 0.168                         & \cellcolor{top3}0.19 & 17.37 & 0.525 & 0.432     \\
3D-GS                               & \cellcolor{top2}332              & 21.56          & 0.847          & 0.130                                 &      \cellcolor{top2}316 & 17.83 & 0.547 & 0.385            \\ \hline
DietNeRF                             & 0.14               & 22.50          & 0.823          & 0.124                               & 0.16 & 17.67 & 0.546 & 0.403          \\
RegNeRF                              & \cellcolor{top3}0.22               & 23.86          & 0.852          & \cellcolor{top3}0.105            & \cellcolor{top3}0.19 & 18.10 & 0.574 & 0.378      \\
FreeNeRF                             & \cellcolor{top3}0.22               & \cellcolor{top2}24.26          & \cellcolor{top2}0.883          & \cellcolor{top2}0.098   &    \cellcolor{top3}0.19 & \cellcolor{top3}18.65 & \cellcolor{top3}0.586 & \cellcolor{top3}0.360            \\
SparseNeRF                           & \cellcolor{top3}0.22               & \cellcolor{top3}24.04          & \cellcolor{top3}0.876          & 0.113              &      \cellcolor{top3}0.19 & \cellcolor{top2}18.81 & \cellcolor{top2}0.591 & \cellcolor{top2}0.354             \\

Ours                              & \cellcolor{top1}467             & \cellcolor{top1}24.64          & \cellcolor{top1}0.895          & \cellcolor{top1}0.095                            & \cellcolor{top1}341 & \cellcolor{top1}19.63 & \cellcolor{top1}0.612 & \cellcolor{top1}0.327        \\  \bottomrule
\end{tabular}

\label{tab:blender}
\end{table}

\begin{figure*}[h]
\vspace{-1mm}
\vskip 0.2in
\begin{center}
    \centering
    \includegraphics[width=0.98\linewidth]{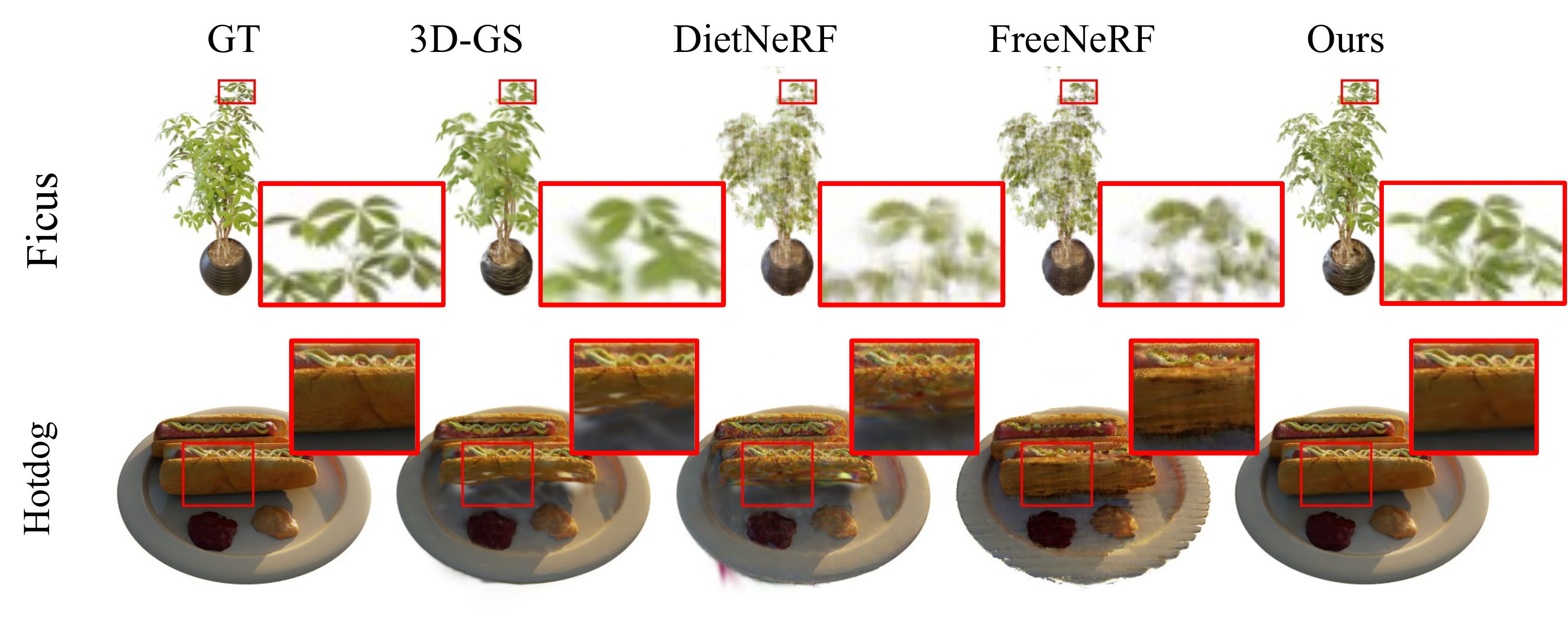}
\vspace{-3mm}
\caption{\textbf{Qualitative Results on Blender Datasets.} Our method consistently outperforms other baselines in the task of novel view synthesis for object-centric datasets. }
\label{fig:exp_blender}
\end{center}
\vspace{-3mm}
\end{figure*}

\paragraph{\textbf{Comparisons on Blender Datasets.}}
The left column of Tab.~\ref{tab:blender} presents the quantitative results on the Blender datasets.
Here, our method significantly outperforms the baselines on object-level datasets, with an improvement of 0.40 in PSNR compared to FreeNeRF, although primarily designed for scene-level scenarios with complex geometry. 
Fig.~\ref{fig:exp_blender} visualizes the rendered image.
We find that DietNeRF hallucinates geometric details,  and FreeNeRF exhibits noticeable aliasing effects. 3D-GS falls short into the excessive blurriness and distorts the edges of the objects. In contrast, our model  not only captures the precise geometry of objects but also accurately simulates their shading effects.

\paragraph{\textbf{Comparisons on Shiny Datasets.}}
We report the quantitative results of the Shiny datasets on the right column of Tab.~\ref{tab:blender}, which feature complex view-dependent effects. Our method performs better than other baselines and improves the PSNR by 1.80dB compared to 3D-GS. The superior performance validates the robustness of our method to handle unusual and challenging materials, such as CDs or glass.

\begin{figure}
    \centering
    \includegraphics[width=0.96\textwidth]{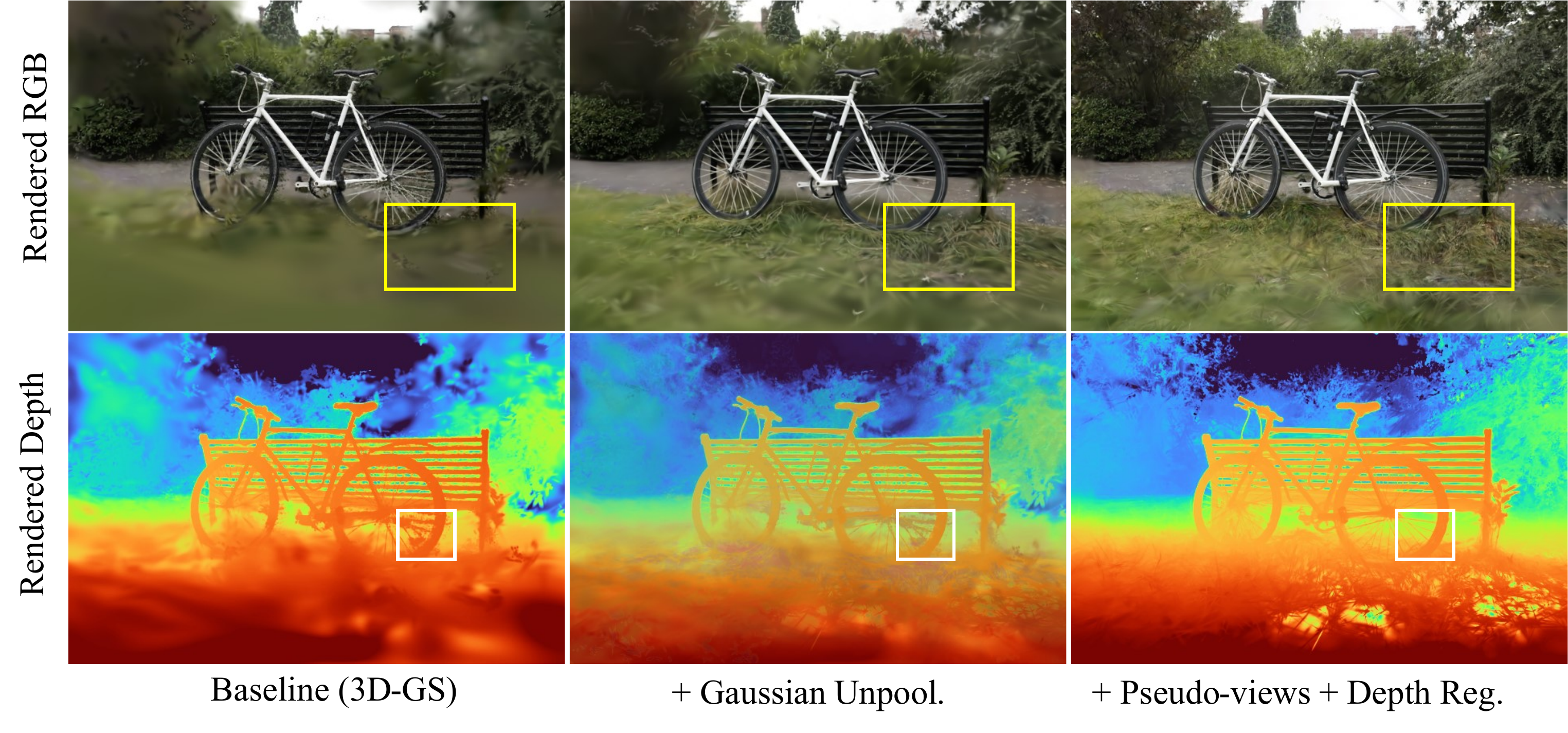}
\vspace{-4mm}
  \caption{\textbf{Ablation Study by Visualization.} 3D-GS~\cite{kerbl20233d} (1st column) shows that the baseline method is significantly degraded when the view coverage is insufficient. \textit{Gaussian Unpooling} provides extra capacity to 3D Gaussians to model the scene, but the learned geometry may not be accurate (2nd column). Adding \textit{Relative Depth Correspondence} regularization (3rd column) can further improve the modeled details.}
    \label{fig:exp_ablation}
\end{figure}

\begin{table}[t]
\small
\caption{{\bf Ablation Study on proposed components.} Starting from 3D-GS~\cite{kerbl20233d} (1st row), we find that our proposed \textit{Gaussian Unpooling} (2nd row) is more effective than the densification scheme in 3D-GS for few-shot view synthesis.
Applying additional supervision from a monocular depth estimator further regularizes the Gaussian optimization towards a better solution (3rd row). Introducing pseudo-view augmentation to apply additional regularization when optimizing Gaussians further enhances the results in a few-shot scenario.
}
\vspace{-3.5mm}
\centering
\resizebox{.63\columnwidth}{!}{
\begin{tabular}{@{}cccccc@{}}
\toprule
 Gaussian & Geometry   & Pseudo   
 & \multirow{2}{*}{PSNR$\uparrow$}  & \multirow{2}{*}{SSIM$\uparrow$}  & \multirow{2}{*}{LPIPS$\downarrow$}  \\
Unpooling & Guidance  & Views  &   &    \\  
  \midrule
\xmark & \xmark & \xmark  &  17.43     &0.522    &0.321 \\
\cmark & \xmark & \xmark  &  18.64     &0.580    &0.311 \\
\cmark & \cmark & \xmark  &  19.83     &0.634    &0.297 \\
\cmark & \cmark & \cmark  &  \textbf{20.31}     & \textbf{0.652}    & \textbf{0.288}\\
 \bottomrule
\end{tabular}}
\vspace{0.4em}
\label{table:ablation_main}
\end{table}

\begin{table}[t]
\caption{\textbf{Ablation Study on different depth estimators.} We utilize various monocular depth estimators on FSGS, and find that FSGS demonstrates strong robustness over different pretrained depth estimators.
}
\vspace{-3.5mm}
\centering
\resizebox{0.64\columnwidth}{!}{
\begin{tabular}{c|cccc} \toprule
\textbf{Method}                     & FPS$\downarrow$ & PSNR$\uparrow$ & SSIM$\uparrow$ & LPIPS$\downarrow$ \\ \hline

SparseNeRF                             & 0.21               & 19.86          & 0.624          & 0.328             \\
3D-GS                               & 385              & 17.43          & 0.522         & 0.321              \\ 
 FSGS (MiDaS small)         & 446                       & 20.17                  & 0.647          & 0.294           \\
 FSGS (DPT Hybrid)           & 460                       & 20.29                  &   0.652        & 0.290  \\
 FSGS (DPT Large)           & 458                       & 20.31                  & 0.652          & \textbf{0.288}  \\
 FSGS (DepthAnything) & \textbf{468}                     & \textbf{20.37}               & \textbf{0.654}         & 0.289               \\ \bottomrule
\end{tabular}}
\label{tab:depth}
\vspace{-2mm}
\end{table}

\subsection{Ablation Studies}
We ablate our design choices on the LLFF dataset under the 3-view setting.
\vspace{-3mm}
\paragraph{\textbf{Effectiveness of Promity-guided Gaussian Unpooling.}}
As shown in the second row of Tab.~\ref{table:ablation_main},  our Promity-guided Gaussian Unpooling expands the scene geometry caused by limited training views, resulting in a PSNR improvement of 1.21dB compared to 3D-GS. We also visualize its visual effects in Fig.~\ref{fig:exp_ablation}. The heuristic Gaussian densification leads to blurring results, particularly noticeable in areas like bush and grass, our approach enriches structural and visual details.
\vspace{-7mm}
\paragraph{\textbf{Impact of Relative Depth Regularization.}}
The third row of Tab.~\ref{table:ablation_main} demonstrates the improvement by introducing depth priors, guiding the Gaussian unpooling towards more plausible geometry. In Fig.~\ref{fig:exp_ablation}, we observe that the depth regularization effectively eliminates the artifacts in grassy regions, and enforces more consistent and solid surfaces with geometric coherence. 
We also display the rendered depth map, where depth regularization leads to depths aligning better with the actual geometric structures.

\vspace{-3mm}
\paragraph{\textbf{Pseudo-view Matters in Few-shot Modeling.}}
Tab.~\ref{table:ablation_main} validates the impact of synthesizing more unseen views during training, which anchors the Gaussians to a plausible geometry and further enhances the modeling quality when the geometry in densification is not accurate.
\vspace{-3mm}
\paragraph{\textbf{Robustness on different pretrained depth estimators.}}
We have shown that FSGS can generalize across various datasets: Blender datasets (object-level), LLFF datasets (indoor), MipNeRF-360 datasets (indoor and outdoor), and self-collected scenes using a mobile phone (Fig.B in the supplementary). To further substantiate its robustness, we employ different monocular depth estimators on LLFF dataset. 
As shown in Tab.~\ref{tab:depth},
all of the depth estimators outperform the baselines, and the Depth-Anything~\cite{depthanything} method achieves the most comparable results.
Collectively, FSGS consistently exhibits strong robustness across different pretrained depth estimators.

\section{Conclusion and Limitation}
In this work, we present a real-time few-shot framework, FSGS, for novel views synthesis within an insufficiently view overlapping.
Starting from extremely sparse point clouds, FSGS adopts the point-based representation and proposes an effective \textit{Proximity-guided Gaussian Unpooling} by measuring the proximity of each Gaussian to its neighbors. The overfitting issue in few-view 3D-GS can be alleviated by the adoption of pseudo-view generation and monocular relative depth correspondences to guide the expanded scene geometry toward a better solution.
FSGS is capable of generating photo-realistic images with as few as three images, and perform inference at more than 200FPS, offering new avenues for real-time rendering and more cost-effective capture methods.

Although FSGS notably enhances the quality and efficiency of real-time few-shot neural rendering, it cannot generalize to occluded views that are unobserved during training. We hope that our proposed approach drives new research towards few-shot novel view synthesis in arbitrary 3D scenes.

\section*{Acknowledgement}
The work is supported by the Intelligence Advanced Research Projects Activity (IARPA) via Department of Interior/Interior Business Center (DOI/IBC) contract number 140D0423C0074. The U.S. Government is authorized to reproduce and distribute reprints for Governmental purposes notwithstanding any copyright annotation thereon. Disclaimer: The views and conclusions contained herein are those of the authors and should not be interpreted as necessarily representing the official policies or endorsements, either expressed or implied, of IARPA, DOI/IBC, or the U.S. Government.

\bibliographystyle{splncs04}
\bibliography{main}

\clearpage
\section{More Technical Details} \label{sec:more_technical}
\subsection{Initialization} 
Similar to 3D Gaussian Splatting~\cite{kerbl20233d}, we start our pipeline from unstructured multi-view images, and calibrate the images using Structure-from-Motion~\cite{schonberger2016structure}. Next, we will continue the dense stereo matching under COLMAP with the function ``patch\_match\_stereo '' and utilize the fused stereo point cloud from ``stereo\_fusion ''. 
We then initialize the SH coefficients at degree 0 and the positions of the 3D Gaussians based on the fused point cloud. Additionally, we remain the rest coefficients and rotation to 0. We also initialize the opacity to 0.1 and set the scale to match the average distance between points.

\subsection{Training}
During training, we start with a SH degree of 0 for a basic lighting representation, incrementing by 1 every 500 iterations up to a degree of 4 to increase complexity over time.
We set the learning rate of position, SH coefficients, opacity, scaling, and rotation to 0.00016, 0.0025, 0.05, 0.005, and 0.001 respectively.
At iterations 2000, 5000, and 7000, the opacity for all Gaussians is reset to 0.05 to eliminate the low-opacity floaters.
In the Blender dataset~\cite{mildenhall2021nerf}, the Pearson correlation is only computed in pixels where the depth values are greater than 0.
Additionally, we utilize an open-source code\footnote{\url{https://pytorch.org/hub/intelisl_midas_v2/}} to estimate the inverse depth map for both the input images and the rendered images from pseudo views.
We detail the procedures of FSGS in Algorithm~\ref{alg:pipeline}.

\begin{table*}
\caption{\textbf{Quantitative Analysis on the Effects of Training Views.} We conduct experiments by using different training views (from 3 to 9) to test the adopted baseline methods. Our FSGS consistently outperforms other methods across all metrics. }
\vspace{-0.5em}
\centering
\begin{tabular}{c|ccc|ccc|ccc} \toprule
\multirow{2}{*}{\textbf{Methods}}    & \multicolumn{3}{c|}{PSNR}                                           & \multicolumn{3}{c|}{SSIM}                                           & \multicolumn{3}{c}{LPIPS} \\ 
 \cline{2-10} 
   & 3-view & 6-view & 9-view 
 & 3-view   & 6-view  & 9-view                
 & 3-view   & 6-view & 9-view                \\ \hline
Mip-NeRF         & 16.10                    & 22.91 & 24.88              & 0.401                    & 0.756 & 0.826    & 0.460 & 0.213 & \cellcolor{top3}0.160     \\
3D-GS        & 17.43                    & 22.87                   & 24.65                     & 0.522                   & 0.732                    & 0.813                     &         \cellcolor{top3}0.321             &  0.204                    &           0.169            \\ \hline

DietNeRF                   & 14.94 & 21.75 &24.28                       & 0.370 & 0.717 & 0.801                     &     0.496 &0.248& 0.183                    \\
RegNeRF           & 19.08                     &23.10 &24.86                  & 0.587 &0.760& 0.820                      & 0.336 & 0.206& 0.161                              \\
FreeNeRF          & \cellcolor{top3}19.63 & \cellcolor{top2}23.73 & \cellcolor{top2}25.13                    & \cellcolor{top3}0.612 &\cellcolor{top3}0.779& \cellcolor{top3}0.827                    & \cellcolor{top2}0.308  & \cellcolor{top2}0.195 & 0.160                     \\
SparseNeRF        & \cellcolor{top2}19.86                     & \cellcolor{top3}23.64                   & \cellcolor{top3}24.97                    & \cellcolor{top2}0.624                    & \cellcolor{top2}0.784                   & \cellcolor{top2}0.834                     &   0.328                         &      \cellcolor{top3}0.202                &  \cellcolor{top2}0.158       \\
Ours         & \cellcolor{top1}20.31                   & \cellcolor{top1}24.55                & \cellcolor{top1}25.89                     &\cellcolor{top1}0.652                   & \cellcolor{top1}0.795                   & \cellcolor{top1}0.845                     &     \cellcolor{top1}0.288                 &   \cellcolor{top1}0.177                   &  \cellcolor{top1}0.143      \\ \bottomrule
\end{tabular}

\label{tab:views}
\vspace{1em}
\end{table*}

\begin{algorithm}[H]
\caption{The training pipeline of FSGS }
\label{alg:pipeline}
\renewcommand{\algorithmicensure}{\textbf{Initialize:}}
\begin{algorithmic}[1]
\State Training view images $\Set{I} = \{I_i \in \real^{H \times W \times 3} \}_{i=1}^{N}$ and their associated camera poses $\Set{P} = \{\Mat{\phi}_i \in \real^{3 \times 4}\}_{i=1}^{N}$. 

\State Run SfM with the input images and camera poses and obtain an initial point cloud $\mathcal{P}$, used to define 3D Gaussians function $\mathcal{G}  = \{G_i(\mu_i, \sigma_i, c_i, \alpha_i) \}_{i=1}^{K}$.

\State Leverage pretrained depth estimator $\mathcal{E}$  to predict the depth map $ D_i = \mathcal{E}(I_i)$.
\State Synthesize pseudo views $\Set{{P^\dag}} = \{\Mat{\phi^\dag}_i \in \real^{3 \times 4}\}_{i=1}^{M}$ from input camera poses  $\Set{P}$.
\While{until convergence}
\State Randomly sample an image $I_i\in \Set{I}$ and the corresponding  camera  pose $\Mat{\phi}_i$
\State  Rasterize the rgb image $\hat{I}_i$  and the depth map $\hat{D}_i$  with camera pose $\Mat{\phi}_i$

\State  $\mathcal{L} = \lambda_1 \lVert I_i - \hat{I}_i \rVert_1 + 
\lambda_2 \operatorname{D-SSIM}(I_i, \hat{I}_i) + 
\lambda_3 \operatorname{Pearson}(D_i, \hat{D}_i)$

\If{$\text{iteration} > t_{iter}$} 
    \State Sample a pseudo camera  pose $\Mat{\phi^\dag}_j \in \Set{P^\dag}$.
    \State  Rasterize the rgb image $\hat{I^\dag}_j$  and the depth  $\hat{D^\dag}_i$ 
    \State Compute the estimated depth as ${D^\dag}_j = \mathcal{E}(\hat{I^\dag}_j)$.
    \State $\mathcal{L} = \mathcal{L} + \lambda_4 \operatorname{Pearson}({D^\dag}_j, \hat{D^\dag}_i)$
\EndIf

\If{IsRefinement(iteration)} 
    \For{$G_i(\mu_i, \sigma_i, c_i, \alpha_i) \in \mathcal{G}$ }
        \If{$\alpha_i > \varepsilon$ or IsTooLarge($\mu_i, \sigma_i$)} 
            \State RemoveGaussian()
        \EndIf 
        \If{$\nabla_{p} \mathcal{L} > t_{pos}$} 
            \State GaussianDensify()
        \EndIf 
        \If{NoProximity($\mathcal{G}$)} 
            \State GaussianUnpooling()
        \EndIf 
    \EndFor
    
\EndIf 

\State Update Gaussians parameter $\mathcal{G}$ via $\nabla_{\mathcal{G}} \mathcal{L}$.
\EndWhile
\end{algorithmic}
\end{algorithm}

\section{More Experiment Results} \label{sec:more_results}
\subsection{Effects of Training Views}
We demonstrate the quantitative results of FSGS on LLFF datasets under 3, 6, 9 views in Tab.~\ref{tab:views}. We can observe that training with more views often leads to better photo-realistic performance  on sparse training data. 
More views provide a more comprehensive coverage of the scene, capturing more details of the scenes. This richness in data boosts supervision signals during optimization, leading to more detailed and structural texture.
Across all the settings, FSGS delivers superior performance compared to all the baselines, affirming the effectiveness of our proposed method.

\begin{figure*}[t]
\vskip 0.2in
\caption{\textbf{Qualitative Results on Datasets Collected by Mobile Phones.} We test the generalization capacity of all methods on self-captured iPhone images, with a calibration process from COLMAP. Our method reveals the majority of scene details despite only adopting three views in training. }
\vspace{-2.2em}
\begin{center}
    \centering
    \includegraphics[width=0.99\linewidth]{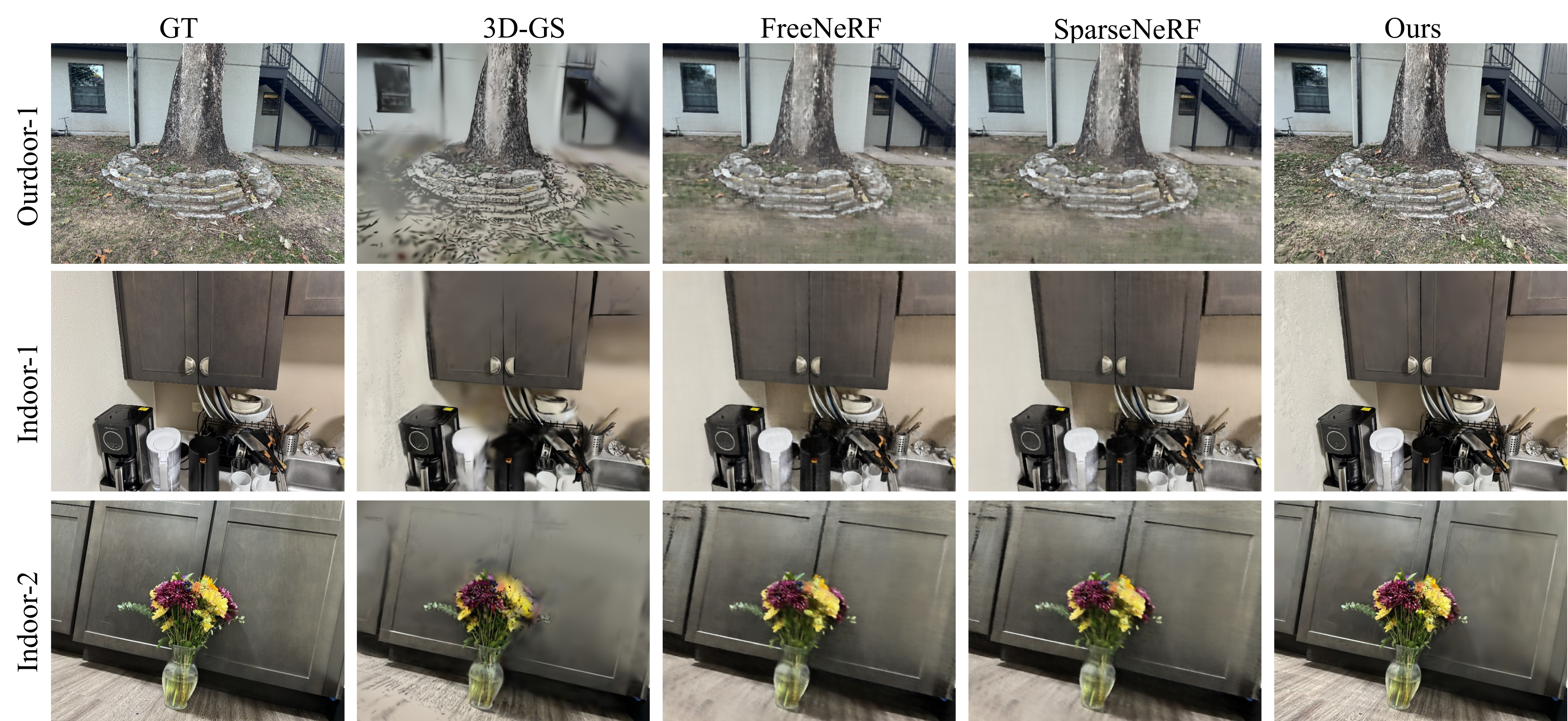}

\label{fig:iphone}
\end{center}
\end{figure*}
\vspace{0.5em}

\begin{table*}[!]
\vspace{4mm}
\caption{\textbf{Quantitative Comparison in Mobile Phone Datasets, with 3 Training Views.}  Our method continues to achieve the best performance in the challenging mobile phone dataset.}
\centering
\begin{tabular}{c|cccc|cccc} \toprule
\multirow{2}{*}{\textbf{Methods}} & \multicolumn{4}{c|}{1/8 Resolution}                                                & \multicolumn{4}{c}{1/4 Resolution}                                                  \\ \cline{2-9}
                                  & FPS$\uparrow$ & PSNR$\uparrow$ & SSIM$\uparrow$ & LPIPS$\downarrow$ & FPS$\uparrow$ & PSNR$\uparrow$ & SSIM$\uparrow$ & LPIPS$\downarrow$ \\ \hline
Mip-NeRF & \cellcolor{top3}0.07 & 14.74 & 0.337 & 0.602 & \cellcolor{top3}0.14 & 13.84 & 0.284 & 0.631 \\
3D-GS & \cellcolor{top2}225 & 16.29 & 0.408 & \cellcolor{top3}0.439 & \cellcolor{top2}127 & 15.83 & 0.413 & 0.530 \\
 \hline
DietNeRF & 0.05 & 13.62 & 0.263 & 0.598 & 0.08 & 12.57 & 0.228 & 0.722 \\
RegNeRF & \cellcolor{top3}0.07 & 17.41 & \cellcolor{top3}0.517 & 0.440 & \cellcolor{top3}0.14 & 16.44 & \cellcolor{top2}0.443 & \cellcolor{top2}0.504 \\
FreeNeRF & \cellcolor{top3}0.07 & \cellcolor{top3}18.07 & 0.497 & \cellcolor{top2}0.426 & \cellcolor{top3}0.14 & \cellcolor{top3}17.14 & \cellcolor{top3}0.462 & \cellcolor{top3}0.509 \\
SparseNeRF & \cellcolor{top3}0.07 & \cellcolor{top2}18.79 & \cellcolor{top2}0.539 & 0.441 & \cellcolor{top3}0.04 & \cellcolor{top2}17.82 & 0.472 & 0.524 \\
Ours & \cellcolor{top1} 263 & \cellcolor{top1}19.54 & \cellcolor{top1}0.539 & \cellcolor{top1}0.403 & \cellcolor{top1} 190 & \cellcolor{top1}18.41 & \cellcolor{top1}0.493 & \cellcolor{top1}0.471 \\
\bottomrule
\end{tabular}
\vspace{3mm}

\label{tab:iphone}
\end{table*}

\subsection{FSGS on Mobile Phones Data}
To validate the generalization capability of FSGS in various real-world settings, we created a new dataset using only a consumer smartphone, the iPhone 15 Pro.
This dataset contains three scenes, comprising two indoor scenes and one outdoor scene. Each scene consists of a collection of RGB images under 5712$\times$4284 resolution, with the viewpoint number ranging from 20 to 40.
Our data calibration pipeline follows the same process procedures as the LLFF datasets~\cite{mildenhall2019local}, and we also select every 8-th image as the novel views for evaluation. For training, we evenly sample 3 images from the remaining views.
These images are then downsampled to 4$\times$ and 8$\times$ for both training and evaluation. We utilize the pretrained monocular depth estimator to predict the depth for input images, and the poses are computed via COLMAP. We compare our method with FreeNeRF~\cite{yang2023freenerf}, SparseNeRF~\cite{wang2023sparsenerf}, and 3D-GS~\cite{kerbl20233d}.

Tab.~\ref{tab:iphone} presents the quantitative results, where FSGS outperforms SparseNeRF with over 0.75 higher PSNR and runs 3,757× faster, a significant leap that underscores its potential for practical, real-world applications where speed is crucial and the environment is intricate.
We also visualize the qualitative results in Fig.~\ref{fig:iphone}, where FSGS significantly improves the visual quality of the scenes over 3D-GS, particularly in the realm of geometry reconstruction. FreeNeRF and SparseNeRF are constrained by geometric continuity from unseen perspectives and do not fully capitalize on the available depth information.

\begin{figure*}[t]
\vskip 0.2in
\begin{center}
    \centering
    \includegraphics[width=0.99\linewidth]{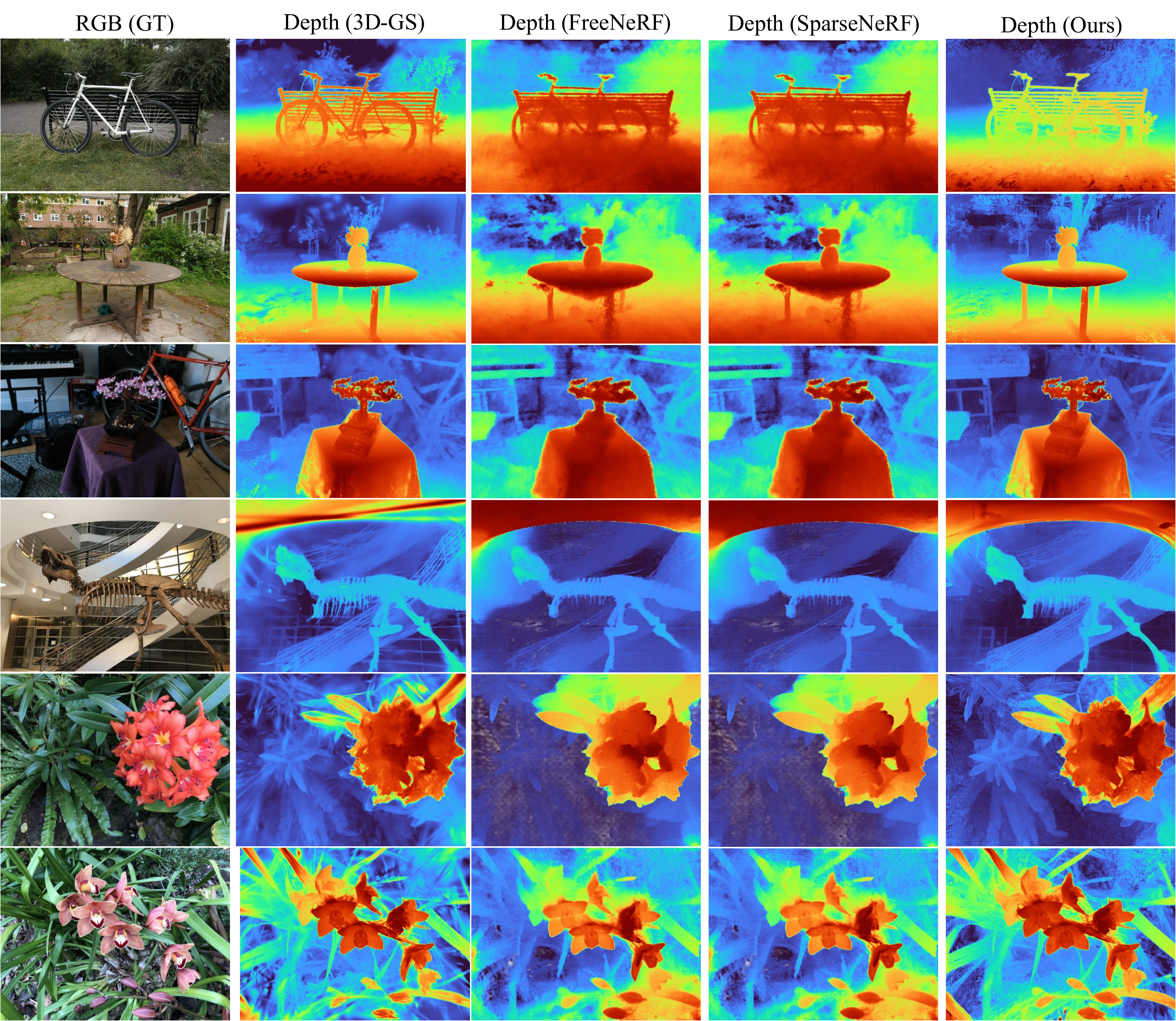}
\vspace{-0.8em}
\caption{\textbf{Visual Comparisons of Predicted Depth.} We visualize the estimated scene depth from all baselines. Noticeable NeRF alias artifacts are found in SparseNeRF and FreeNeRF. 3D-GS produces an oversmoothed scene geometry, while our method demonstrates visually pleasing geometric details. }
\label{fig:depth}
\end{center}
\end{figure*}

\vspace{7mm}
\subsection{Visual Comparisons of the Rendered Depth}
\vspace{3mm}
We demonstrate the qualitative results of the predicted depth for each methods, as shown in Fig.~\ref{fig:depth}. We compare our method with 3DGS~\cite{kerbl20233d}, FreeNeRF~\cite{yang2023freenerf} and SparseNeRF~\cite{wang2023sparsenerf}. In the left we visualize the  ground truth of the images. 
FSGS significantly outperforms the three baselines in terms of depth quality and details. 
The depth maps produced by FSGS are not only more accurate but also exhibit a higher level of detail, showcasing its robustness in reconstructing complex scenes. 
In contrast, both FreeNeRF and SparseNeRF exhibit limitations in geometric modeling and struggle to accurately learn complex geometries, leading to a distorted scene representation. Although SparseNeRF leverage depth prior, it still does not fully capture the fine-grained structures in real-world structures, resulting in a noticeable drop in quality compared to FSGS.
3D-GS, on the other hand, tends to lose fine details in the areas from away the camera, leading to a diminished overall quality in the depth and texture of distant objects.

\end{document}